\pdfoutput=1

\documentclass[11pt]{article}

\usepackage[final]{acl}

\usepackage{times}
\usepackage{latexsym}
\usepackage{amsmath}
\usepackage{multirow}
\usepackage{pifont}

\usepackage[T1]{fontenc}

\usepackage[utf8]{inputenc}

\usepackage{microtype}

\usepackage{inconsolata}

\usepackage{arydshln}
\usepackage{makecell}
\usepackage{tcolorbox}
\tcbuselibrary{breakable}

\definecolor{darkgreen}{rgb}{0.0, 0.5, 0.0}
\title{TinyThinker: Distilling Reasoning through Coarse-to-Fine Knowledge Internalization with Self-Reflection}

\author{Shengmin Piao \\
  Yonsei University \\
  Seoul, South Korea \\
  \texttt{shengminp@yonsei.ac.kr} \\
  \And
  Sanghyun Park\textsuperscript{\textdagger} \\
  Yonsei University \\
  Seoul, South Korea \\
  \texttt{sanghyun@yonsei.ac.kr} \\}

\begin{document}
\maketitle
\renewcommand{\thefootnote}{}
\footnotetext{\textsuperscript{\textdagger} Corresponding author.}
\renewcommand{\thefootnote}{\arabic{footnote}}
\begin{abstract}
Large Language Models exhibit impressive reasoning capabilities across diverse tasks, motivating efforts to distill these capabilities into smaller models through generated reasoning data. However, direct training on such synthesized reasoning data may lead to superficial imitation of reasoning process, rather than fostering a genuine integration of reasoning capabilities with underlying knowledge. To address this, we propose TinyThinker, a framework introducing two novel approaches. First, we introduce a three-stage process that incrementally guides the student model through the reasoning process, progressively refining knowledge from coarse to fine granularity. Second, we develop a two-phase training framework comprising an initial reasoning acquisition phase followed by a self-reflection phase utilizing self-generated data. Experiments on commonsense reasoning benchmarks demonstrate that TinyThinker achieves superior performance compared to baselines. Ablation studies further validate the effectiveness of each component in our framework. We expect that TinyThinker can be extended to other knowledge-intensive reasoning tasks, offering an alternative strategy for developing effective reasoning capabilities in smaller language models. Codes are available at \href{https://github.com/shengminp/TinyThinker}{https://github.com/shengminp/TinyThinker}.
\end{abstract}

\section{Introduction}
Large language models (LLMs) demonstrate impressive reasoning capabilities across a wide range of tasks \citep{kaplan2020scaling, wei2022emergent}, largely facilitated by \textit{in-context learning} \citep{NEURIPS2020_1457c0d6} and Chain-of-Thought (CoT) prompting \citep{wei2022chain}. These methods enable LLMs to generalize from examples of intermediate reasoning steps during inference, enabling them to effectively address complex multi-step reasoning tasks, particularly in commonsense and arithmetic reasoning domains \citep{achiam2023gpt}.

\begin{figure}[t]
  \includegraphics[width=\columnwidth]{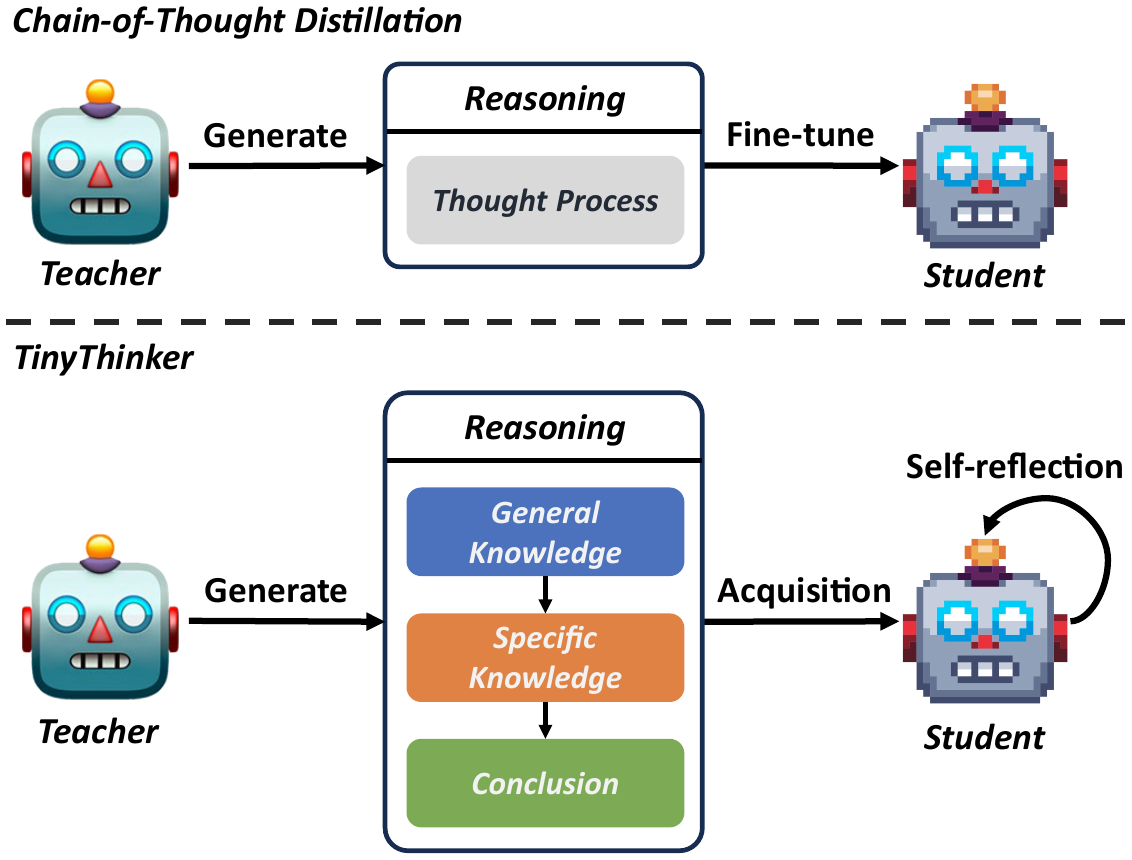}
  \caption{Comparison between TinyThinker and standard Chain-of-Thought Distillation. \textbf{Top:} Fine-tuning the student directly on teacher-generated reasoning data. \textbf{Bottom:} TinyThinker acquires reasoning capabilities through a three-stage process, further refined via self-reflection.}
  \label{fig:overall framework}
\end{figure}

A notable limitation of CoT prompting is its reliance on models with over 100 billion parameters, making it computationally expensive and less feasible for real-world deployment \citep{wei2022chain}. To address these limitations, prior works \citep{hsieh-etal-2023-distilling, ho-etal-2023-large, magister-etal-2023-teaching} have explored leveraging LLMs to generate reasoning data for knowledge distillation, thereby strengthening the reasoning capabilities of smaller language models (LMs) \footnote{In this study, "large" refers to models with over 100 billion parameters, whereas "small" denotes those with fewer than 1 billion parameters.}. Building on these advances, recent studies generally follow two main steps: (1) applying CoT prompting or its variants \citep{kojima2022large} to generate reasoning data from a teacher model, and (2) designing specialized fine-tuning strategies to transfer reasoning capabilities to a student model\footnote{In accordance with knowledge distillation terminology, LLMs are refereed to as teacher models, and smaller LMs as student models.} \citep{shridhar-etal-2023-distilling, jiang-etal-2023-lion, kang2024knowledge}.

Although these methods have yielded promising results, it remains uncertain whether the student model genuinely develops reasoning capabilities. A recent study attributes the success of CoT prompting to its decomposition of compositional functions during in-context learning, enabling the model to focus on relevant data at each step and learn single-step composition functions in context \citep{li2024dissecting}. This interpretation aligns with the perspective that LLMs perform multi-step reasoning internally \citep{hou-etal-2023-towards}. However, when the student model lacks such intrinsic capabilities, direct training on synthesized reasoning data raises a critical concern: the student model may imitate the style of the synthesized data rather than its factuality \citep{gudibande2024the}. Given that reasoning typically involves multiple inference steps—each requiring explicit or implicit application of knowledge \citep{yu2023natural}—there is a risk that the student model might mimic these step-by-step reasoning processes superficially, without truly internalizing the underlying knowledge.

To enable flexible incorporation of internal knowledge for effective reasoning, we propose \textit{TinyThinker}, a novel framework that structures the reasoning process into three stages, progressively refining knowledge from coarse to fine granularity (Figure~\ref{fig:overall framework}). To facilitate the acquisition and refinement of this structured reasoning process, we introduce a two-phase training framework, consisting of a \textit{reasoning acquisition} phase followed by a \textit{self-reflection} phase.

This study focuses on commonsense reasoning, which inherently requires the integration of extensive knowledge with effective reasoning capabilities. Given that most commonsense reasoning datasets are structured in a multiple-choice question (MCQ) format \citep{talmor-etal-2019-commonsenseqa, mihaylov-etal-2018-suit, geva-etal-2021-aristotle}, we designed a three-stage process inspired by human problem-solving strategies. In the reasoning acquisition phase, the model first \textit{recalls} general knowledge relevant to the question and options. It then conducts an \textit{analysis} of specific knowledge for each option, using the recalled general knowledge as context. Finally, the model integrates the acquired knowledge to \textit{summarize} and identify the correct answer.

After acquiring new reasoning capabilities, independent reflection is essential for consolidating these capabilities. In the self-reflection phase, the student revisits previous training data and generates new reasoning data through the same three-stage process. These self-generated data, combined with iterative Direct Preference Optimization (DPO) \citep{rafailov2024direct}, further refine the reasoning capabilities, which fosters a deeper internalization of the underlying knowledge.

Experimental results on three commonsense reasoning datasets (CommonsenseQA, OpenBookQA, and StrategyQA) demonstrate that TinyThinker significantly enhances the reasoning capabilities of the student model. Furthermore, the scalability of our approach is validated by consistent performance improvements as the model size increases. Additionally, ablation studies on both the three-stage reasoning process and the self-reflection phase further underscore the effectiveness of the coarse-to-fine reasoning approach, along with the model's iterative self-reflection capabilities through the generation of internal knowledge.

\section{Related Work}

\begin{figure*}[t]
  \includegraphics[width=\linewidth]{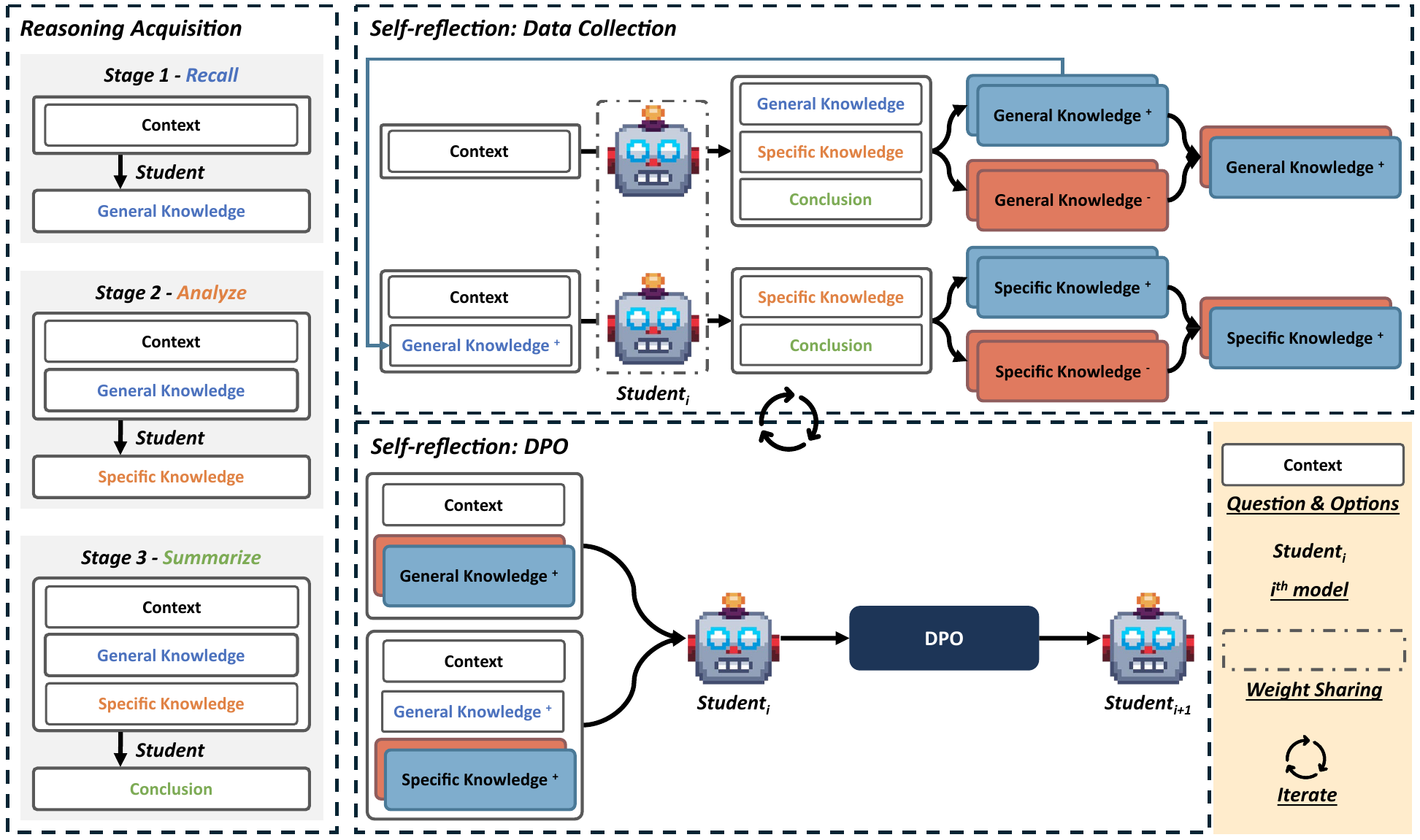}
  \caption {Detailed process of TinyThinker. \textbf{Reasoning Acquisition:} The student model follows a recall-analyze-summarize process, refining reasoning from coarse to fine granularity. \textbf{Self-reflection:} The model iteratively collects data and applies DPO. Pairwise data is first collected during the recall stage, and the preferred data from this stage informs the collection of pairwise data in the analyze stage. Once sufficient data is gathered, DPO is applied to refine the student's reasoning capabilities, facilitating progression to the next iteration of self-reflection.}
  \label{fig:specific framework}
\end{figure*}

Knowledge distillation transfers knowledge from a larger teacher model to a smaller student model \citep{gou2021knowledge}, enabling the student to leverage the features learned by the teacher \citep{hinton2015distilling}. As LLMs advance in generating high-quality data, this process has been extended to transfer reasoning capabilities via teacher-generated data. Recent research in this area can be categorized into five main approaches: supervised fine-tuning, decomposer-solver framework, feedback-based framework, retriever-augmented framework, and self-improvement framework.

\paragraph{Supervised fine-tuning} This method trains the student model exclusively on teacher-generated reasoning data, typically employing a language modeling objective. It has become a standard approach for reasoning distillation \citep{ho-etal-2023-large, magister-etal-2023-teaching, pmlr-v202-fu23d, wang-etal-2023-scott, li-etal-2023-symbolic}. Alternatively, reasoning generation is combined with direct answer prediction within a multi-task learning framework, which has been shown to yield further performance improvements \citep{hsieh-etal-2023-distilling, chen-etal-2024-learning-maximize, li2024explanations}. However, these methods rely solely on teacher-generated data and lack an explicit focus on the integration of reasoning with underlying knowledge.

\paragraph{Decomposer-solver framework} This framework trains two student models: one responsible for decomposing complex questions into simpler sub-questions, and another tasked with solving these sub-questions. The approach can be implemented in a single-step process, where both decomposition and solution occur in one iteration \citep{shridhar-etal-2023-distilling}, or as a turn-based approach, where decomposition and solving are repeated iteratively until a complete solution is reached \citep{han-etal-2023-dialcot}. While this method effectively simplifies complex reasoning questions, it primarily focuses on semantic decomposition.

\paragraph{Feedback-based framework} In this framework, the student model is initially trained to generate reasoning processes. When the student produces an incorrect answer, the associated reasoning process is sent back to the teacher, which provides feedback on the errors. This feedback is incorporated into subsequent rounds of fine-tuning, creating an iterative cycle of corrective guidance that gradually improves the student's reasoning capabilities \citep{jiang-etal-2023-lion, wang-etal-2023-democratizing}. While this approach enables the student to address its deficiencies through targeted feedback, the reliance on the teacher limits the student to independently infer connections between reasoning and underlying knowledge.

\begin{table*}[t]
\renewcommand{\arraystretch}{0.85}
  \centering
  \begin{tabular}{lp{12.4cm}}
    \Xhline{1.0pt}
    \textbf{Stage} & \textbf{Data Format} \\
    \Xhline{0.5pt}
    \small{Data Curation} & \small{\textbf{Prompt:}
                    \texttt{Question Options: (A)... (B)... (C)... (D)...\textbackslash n Key Information: <>\textbackslash n Explanations: A is correct. Because <> B is incorrect. Because <> C is incorrect. Because <> D is incorrect. Because <>}
                    \newline \textit{(Note: "<>" denotes knowledge to be generated by the teacher model.)}} \\
    \Xhline{0.5pt}
    \small{Recall} & \small{\textbf{Input:}
             \texttt{Question Options: (A)... (B)... (C)... (D)...\textbackslash n \textbf{Recall:}}
             \newline \textbf{Label: }\texttt{General Knowledge}} \\
    \Xhline{0.5pt}
    \small{Analyze} & \small{\textbf{Input:}
              \texttt{Question Options: (A)... (B)... (C)... (D)...\textbackslash n Recall: General Knowledge.\textbackslash n \textbf{Analyze: For option A,}}
              \newline \textbf{Label: }\texttt{Specific Knowledge}
              \newline \textit{(Note: Apply the same format for option B, C, D)}} \\
    \Xhline{0.5pt}
    \small{Summarize} & \small{\textbf{Input:}
                \texttt{Question Options: (A)... (B)... (C)... (D)...\textbackslash n Recall: General Knowledge.\textbackslash n Analyze: For option A, Specific Knowledge. For option B, Specific Knowledge. For option C, Specific Knowledge. For option D, Specific Knowledge.\textbackslash n \textbf{Summarize:}}
                \newline \textbf{Label: }\texttt{Summary}} \\
    \Xhline{1.0pt}
  \end{tabular}
  \caption{Data format across various stages of TinyThinker: data curation, recall, analyze, and summarize stages.}
  \label{tab:data_format}
\end{table*}

\paragraph{Retriever-augmented framework} In contrast to the feedback-based framework, which focuses on refining reasoning through direct guidance, the retriever-augmented framework improves reasoning by integrating external information. Specifically, a retriever is employed to search for relevant information from external sources, such as Wikipedia, and incorporated into the reasoning process to improve performance \citep{kang2024knowledge, li-etal-2024-teaching}. Although this framework strengthens the student's reasoning by supplementing it with external information, the retrieved information is not integrated into the student's internal knowledge.

\paragraph{Self-improvement framework} Unlike the feedback-based and retriever-augmented frameworks, which rely on external assistance from either the teacher model or knowledge bases, the self-improvement framework refines the reasoning capabilities through self-generated data \citep{liu-etal-2023-crystal}. In this approach, the student initially learns basic reasoning capabilities from the teacher. It then generates both correct and incorrect relevant knowledge, which is used as paired data for reinforcement learning. While this approach underscores the complementarity between knowledge and reasoning, its primary focus is on enabling the student to generate and utilize its internal knowledge effectively, rather than on improving the reasoning process itself.

Our approach extends the self-improvement framework with two key innovations: (1) a novel three-stage reasoning process that equips the student model with foundational reasoning capabilities by internalizing underlying knowledge, and (2) a self-reflection method that iteratively applies the DPO approach to further refine the relationship between reasoning capabilities and underlying knowledge.

\section{Methodology}
To enable the student to internalize the underlying knowledge for effective reasoning, we propose a two-phase training approach (Figure~\ref{fig:specific framework}): reasoning acquisition (§\ref{subsec:reasoning_acquisition}) and self-reflection (§\ref{subsec:self_reflection}). During the reasoning acquisition phase, the student model progresses through three stages—recall, analyze, and summarize—which progressively refine its reasoning from general to specific knowledge. In the subsequent self-reflection phase, the student generates multiple reasoning processes based on its learned capabilities to further refine the recall and analyze stages with DPO.

\subsection{Reasoning Acquisition}
\label{subsec:reasoning_acquisition}

\paragraph{Data Curation} The initial dataset consists of a question $q$ and options $O = \{o_1, o_2, \dots, o_n\}$. To generate data for each stage, we utilized a few-shot prompting method with 7 to 8 examples, adapted from prior work \citep{wei2022chain, wang2023selfconsistency, li2024explanations} and tailored to our task (Table~\ref{tab:data_format}). Following \citet{magister-etal-2023-teaching}, we explicitly indicate the correctness of each option, encouraging the teacher model to generate accurate knowledge. Full prompt details are provided in Appendix~\ref{sec:full prompts}.

Once sufficient data had been collected, they were reorganized for each stage as shown in Table~\ref{tab:data_format}. Specifically, stage-specific cue phrases such as "Recall:", "Analyze: For Option", and "Summarize:" were introduced to guide the student model, parameterized by $\theta$, in generating general knowledge $k_{\mathrm{general}}$, option-specific knowledge $k_{\mathrm{specific}}$, and a final summary $k_{\mathrm{summary}}$.

\paragraph{Recall Stage} The student performs a cursory scan of the question and options, retrieving general background knowledge. This broad knowledge serves as a foundation for subsequent reasoning processes. The training objective for this stage is expressed as:
\begin{equation}
  \label{eq:recal_stage}
  \mathcal{L_{\mathrm{recall}}}(\theta) = -\log P(k_{\mathrm{general}} \mid q, O; \theta)
\end{equation}

\paragraph{Analyze Stage} After recalling general knowledge, the student carefully evaluates each option $o_i \in O$, generating specific knowledge relevant to each option. Although the model may infer the correctness of certain options, it can be misled by confounding alternatives, necessitating a detailed analysis to reach a reliable summary. The training objective for this stage is formalized as:
\begin{equation}
  \label{eq:analyze_stage}
  \mathcal{L_{\mathrm{analyze}}}(\theta) = -\log P(k_{\mathrm{specific}} \mid q, O, k_{\mathrm{general}}; \theta)
\end{equation}

\paragraph{Summarize Stage} At this stage, all previously generated knowledge is fed back into the model, enabling it to derive a final summary and select the correct answer. The training objective for this stage is defined as:
\begin{equation}
\begin{split}
  \label{eq:summarize_stage}
  &\mathcal{L_{\mathrm{summary}}}(\theta) \\
  &= -\log P(k_{\mathrm{summary}} \mid q, O, k_{\mathrm{general}}, k_{\mathrm{specific}}; \theta)
\end{split}
\end{equation}

\subsection{Self-Reflection}
\label{subsec:self_reflection}
\paragraph{Iterative DPO} Direct Preference Optimization (DPO) aligns models with human preferences using pairwise comparisons $\mathcal{D}=\left\{{x_i, y_i^w, y_i^l}\right\}_{i=1}^{N}$, where $x_i$ is the input to the model $\pi_\theta$, and $y_i^w$ and $y_i^l$ are outputs generated by the model. Human evaluators identify one output as "preferred" ($y^w$) and the other as "dispreferred" ($y^l$).

In this study, we adapt DPO to refine the student's reasoning capabilities by aligning it with its internal knowledge. We employ an iterative DPO approach \citep{pmlr-v235-yuan24d, pang2024iterative} to facilitate the self-reflection process. Specifically, this approach trains a series of models \{$\pi_1, ..., \pi_T$\}, where each model at iteration $t$ is trained on new pairwise data generated by the model from the preceding iteration $({t-1})$. At each iteration, the model $\pi_\theta$ is initialized with the parameters of the $({t-1})^\mathrm{th}$ model. 

To further stabilize training, an additional negative log-likelihood (NLL) loss term is applied to the preferred data, following \citet{pang2024iterative} and \citet{dubey2024llama}. The overall training objective for DPO combined with NLL is defined as:
\begin{equation}
\begin{split}
&\mathcal{L}_{\mathrm{DPO+NLL}}(\pi_t; \pi_\theta) \\
&= \mathcal{L}_{\mathrm{DPO}}(y_i^w, y_i^l | x_i) + \alpha \mathcal{L}_{\mathrm{NLL}}(y_i^w | x_i) \\
&= - \log \sigma \left( \beta \log \frac{\pi_t(y_i^w | x_i)}{\pi_\theta(y_i^w | x_i)} - \beta \log \frac{\pi_t(y_i^l | x_i)}{\pi_\theta(y_i^l | x_i)} \right) \\
&\quad - \alpha \log \pi_t(y_i^w | x_i)
\end{split}
\end{equation}

\paragraph{Data Collection} Unlike standard DPO, which leverages continuous rewards to determine preferred outputs, we adopt a binary approach that classifies the generated outputs as either correct or incorrect based on their corresponding summaries. This assessment directly reflects the accuracy of the model's underlying knowledge: a correct summary indicates the presence of accurate knowledge, which warrants prioritization for further refinement. In contrast, an incorrect summary reveals deficiencies in relevant knowledge and signals the need for improvement.

In our approach, knowledge is generated primarily during the recall and analyze stages. To strengthen each stage individually, we collect data by following the three-stage process twice, with different generation strategies for each pass. In the first data collection, temperature sampling is used in the recall stage to generate diverse knowledge, while greedy search is applied in the analyze and summarize stages to minimize influence on the final summary. Since the analyze stage depends on the outputs from the recall stage, the second data collection starts directly at the analyze stage, utilizing the "preferred" general knowledge gathered from the first collection. Similarly, temperature sampling is applied to the analyze stage, and greedy search is employed for the summarize stage.

\begin{figure}[t]
  \includegraphics[width=\columnwidth]{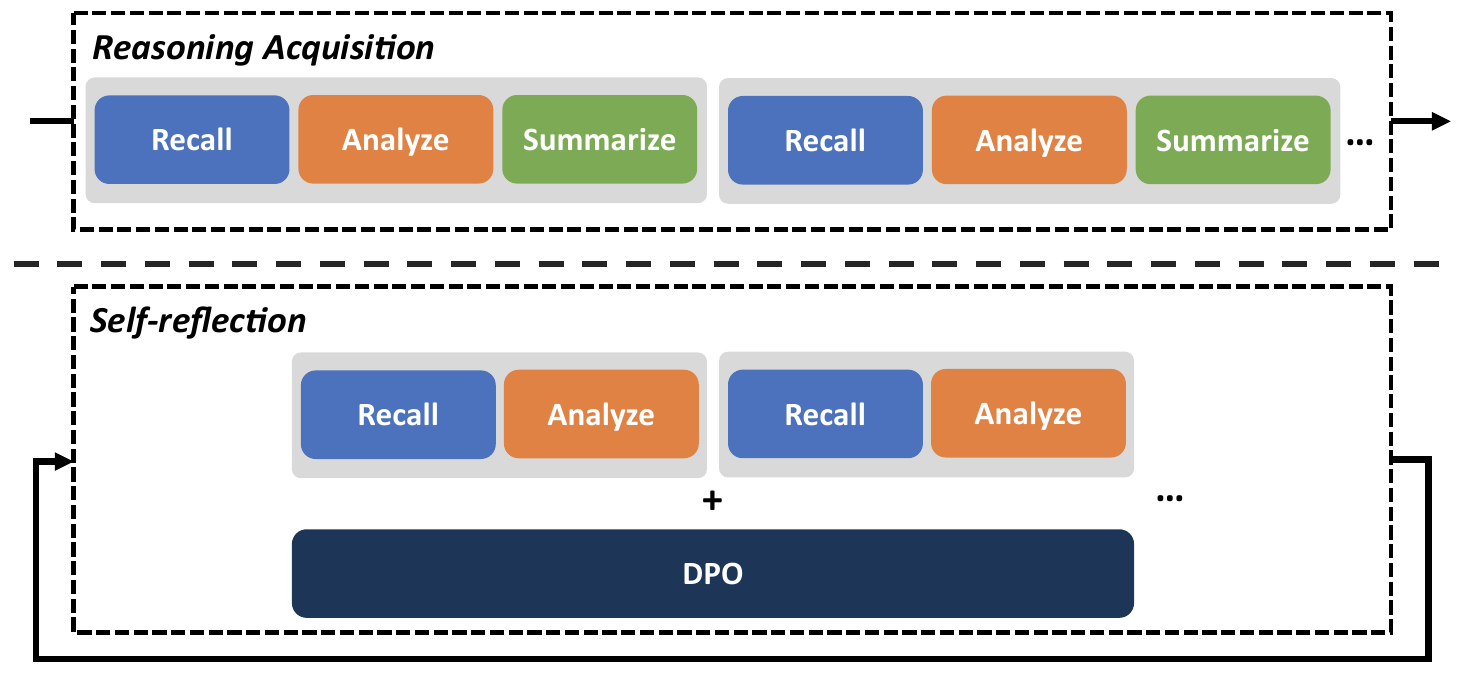}
  \caption{Overall process of the training strategy. \textbf{Top:} During the reasoning acquisition phase, the recall-analyze-summarize process is repeated iteratively. \textbf{Bottom:} During the self-reflection phase, the recall-analyze process is iterated with DPO.}
  \label{fig:training strategy}
\end{figure}

\subsection{Training Strategy} 
To effectively apply the three-stage reasoning process to novel questions, we repeat this process multiple times per epoch during the reasoning acquisition phase (Figure~\ref{fig:training strategy}). Each iteration consists of a predetermined number of training steps per stage, with the model advancing sequentially through the stages. Since the analyze stage requires generating option-specific knowledge, the corresponding dataset is proportionally larger, requiring more training steps compared to the recall and summarize stages.

After learning the three-stage reasoning process, the student transitions into the self-reflection phase to further refine its reasoning capabilities. This phase retains the iterative structure but differs in two key aspects: (1) the summarize stage is excluded since only the recall and analyze stages require further refinement, and (2) the analyze stage now focuses on pairwise comparisons for each option individually, leading to an equal number of training steps for both recall and analyze stages.

\section{Experimental Setup}
\subsection{Dataset}
We evaluate the student model on three commonsense reasoning benchmarks: CommonsenseQA (CSQA) \citep{talmor-etal-2019-commonsenseqa}, OpenBookQA (OBQA) \citep{mihaylov-etal-2018-suit}, and StrategyQA \citep{geva-etal-2021-aristotle}. These benchmarks require substantial knowledge and reasoning capabilities to achieve competitive performance. A detailed description of each dataset is provided in Appendix~\ref{sec:datasets}.

\begin{table*}[t]
    \centering
    \begin{tabular}{llccc}
        \Xhline{1.0pt}
        \textbf{Model} & \textbf{Methods} & \textbf{CSQA} & \textbf{OBQA} & \textbf{StrategyQA}$^a$\\ \Xhline{0.5pt}
        \multirow{7}{*}{\textit{T5-Small (60M)}}
        & Fine-tune-CoT \citep{ho-etal-2023-large} & 29.48 & - & 56.04\\ 
        & DSS \citep{hsieh-etal-2023-distilling} & 43.24 & - & -\\ 
        & MI Distillation \citep{chen-etal-2024-learning-maximize}& 43.90 & - & -\\ 
        & MT-CoT \citep{li2024explanations}& \textbf{49.17} & 51.72 & -\\
        & D\&R Distillation \citep{li-etal-2024-teaching} & - & - & 55.00\\
        \cdashline{2-5}
        & \textbf{TinyThinker} & 46.36 & \textbf{53.60} & \textbf{60.26}\\ 
        \Xhline{0.5pt}
        \multirow{8}{*}{\textit{T5-Base (220M)}}
        & Fine-tune-CoT \citep{ho-etal-2023-large} & 45.37 & - & 59.68\\ 
        & DSS \citep{hsieh-etal-2023-distilling} & 63.29 & - & -\\ 
        & KARD \citep{kang2024knowledge} & - & 59.33 & 56.57\\
        & MI Distillation \citep{chen-etal-2024-learning-maximize} & 63.88 & - & -\\ 
        & MT-CoT \citep{li2024explanations}& \textbf{64.50} & 60.68 & 61.05 \\
        & D\&R Distillation \citep{li-etal-2024-teaching} & - & - & 59.00\\
        \cdashline{2-5}
        & \textbf{TinyThinker} & 59.79 & \textbf{62.40} & \textbf{66.38}\\ 
        \Xhline{0.5pt}
        \multirow{8}{*}{\textit{T5-Large (770M)}}
        & Fine-tune-CoT \citep{ho-etal-2023-large} & 54.22 & - & 62.15\\ 
        & DSS \citep{hsieh-etal-2023-distilling} & 70.43 & - & -\\ 
        & KARD \citep{kang2024knowledge} & - & 66.40 & 66.04\\
        & Crystal \citep{liu-etal-2023-crystal} & 70.52 & 64.20 & -\\ 
        & MT-CoT \citep{li2024explanations}& \textbf{74.37} & 64.60 & -\\
        & D\&R Distillation \citep{li-etal-2024-teaching} & - & - & 63.30\\
        \cdashline{2-5}
        & \textbf{TinyThinker} & 65.44 & \textbf{68.80}& \textbf{69.00}\\ 
        \Xhline{1.0pt}
        \multicolumn{5}{l}{\footnotesize{$^a$ Due to varying data splits in StrategyQA across papers, the results are for reference only.}}
    \end{tabular}
    \caption{Accuracy (\%) of the student model across baselines. \textbf{Bold} values indicate the best performance.}
    \label{tab:overall-performance-part}
\end{table*}

\subsection{Baselines}
We compare the performance of TinyThinker with several existing reasoning distillation approaches that use the same datasets and student model architectures:

\textbf{Fine-tune-CoT} \citep{ho-etal-2023-large} applies standard supervised fine-tuning to train the student model to generate CoT-style reasoning. Building on this, \textbf{DSS} \citep{hsieh-etal-2023-distilling} and \textbf{MT-CoT} \citep{li2024explanations} incorporate multi-task learning to train the student model to generate reasoning while simultaneously predicting the correct answer. \textbf{MI Distillation} \citep{chen-etal-2024-learning-maximize} further improves multi-task learning by maximizing the mutual information between reasoning generation and answer prediction. The retrieval-augmented method \textbf{KARD} trains a retriever to assist the reasoning generation process directly, while \textbf{D\&R Distillation} trains two separate student models: one to decompose questions into subquestions and another to answer these subquestions using an external knowledge base. Finally, \textbf{Crystal} \citep{liu-etal-2023-crystal} generates substantial amounts of relevant knowledge and incorporates it into the reasoning process simultaneously, then applies reinforcement learning to encourage the student model to generate more relevant knowledge.

\subsection{Implementation Details}
\paragraph{Teacher Model} We employ GPT-4o (gpt-4o-2024-05-13) as the teacher model, accessed via the OpenAI API\footnote{https://platform.openai.com/docs/model/gpt-4o}. The teacher model generates reasoning data for each question in the dataset following the three-stage reasoning process as outlined in our prompt design.

\paragraph{Student Model} We utilize the T5 models—Small (60M), Base (220M), and Large (770M) \citep{raffel2020exploring}—as the backbone for the student model. All models are trained using 4 A100 GPUs via the Huggingface library \citep{wolf2019huggingface}, with pretrained weights from publicly available sources\footnote{https://huggingface.co/google-t5}. Detailed hyperparameter settings are provided in Appendix~\ref{sec:hyperparameter}.

\section{Results and Analysis}
\subsection{Overall Performance}
We applied TinyThinker to train all student models across the datasets, with accuracy (\%) as the primary evaluation metric. The performance results, summarized in Table~\ref{tab:overall-performance-part}, are organized by student model size and compared against baseline methods that trained the same student models using different approaches on the same datasets. 

TinyThinker consistently achieves the best performance on both the OBQA and StrategyQA datasets, surpassing the best-performing baseline by margins of 1.5\% to 2\% on OBQA and 3\% to 7\% on StrategyQA across all model sizes. This demonstrates the advantage of internalizing underlying knowledge for effective reasoning. 

Although TinyThinker does not outperform MT-CoT on the CSQA dataset, it outperforms MT-CoT on both the OBQA and StrategyQA datasets. In contrast, TinyThinker maintains a consistent performance gap of approximately 5\% compared to DSS and MI Distillation on the CSQA dataset. Notably, all three methods—MT-CoT, DSS, and MI Distillation—utilize multi-task learning to develop reasoning capabilities. However, directly predicting answers in this approach may lead the student to rely on spurious correlations between question and options, potentially exploiting reasoning shortcut to arrive at the answer \citep{wang-etal-2023-scott, wang2022pinto}.

\subsection{Performance Across Model Sizes}
\begin{figure}[t]
  \includegraphics[width=\linewidth]{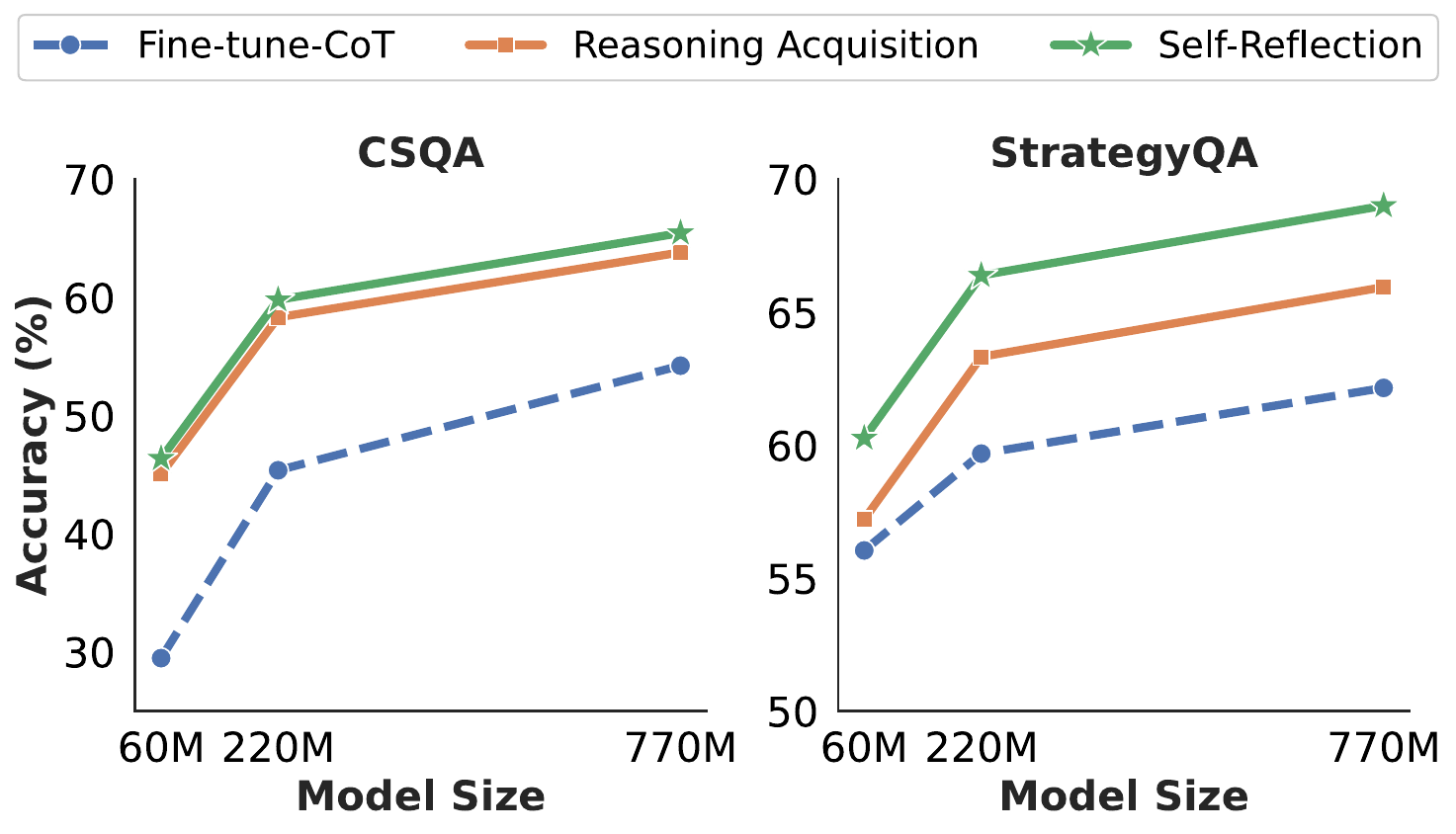}
  \caption {Accuracy (\%) on CSQA and StrategyQA datasets across different model sizes.}
  \label{fig:scalability}
\end{figure}

We evaluated TinyThinker's performance across different model sizes on the CSQA and StrategyQA datasets, comparing it with the Fine-tune-CoT baseline. As illustrated in Figure~\ref{fig:scalability}, TinyThinker consistently outperforms Fine-tune-CoT across all model sizes, indicating the advantages of the proposed three-stage reasoning process over the standard fine-tuning approach. Both the reasoning acquisition and self-reflection phases improve with increasing model size, demonstrating the scalability of our approach. Furthermore, the self-reflection phase consistently enhances the performance of the reasoning acquisition phase, confirming that refinement through self-generated data strengthens the student's reasoning capabilities.

\begin{figure*}[t]
  \includegraphics[width=\linewidth]{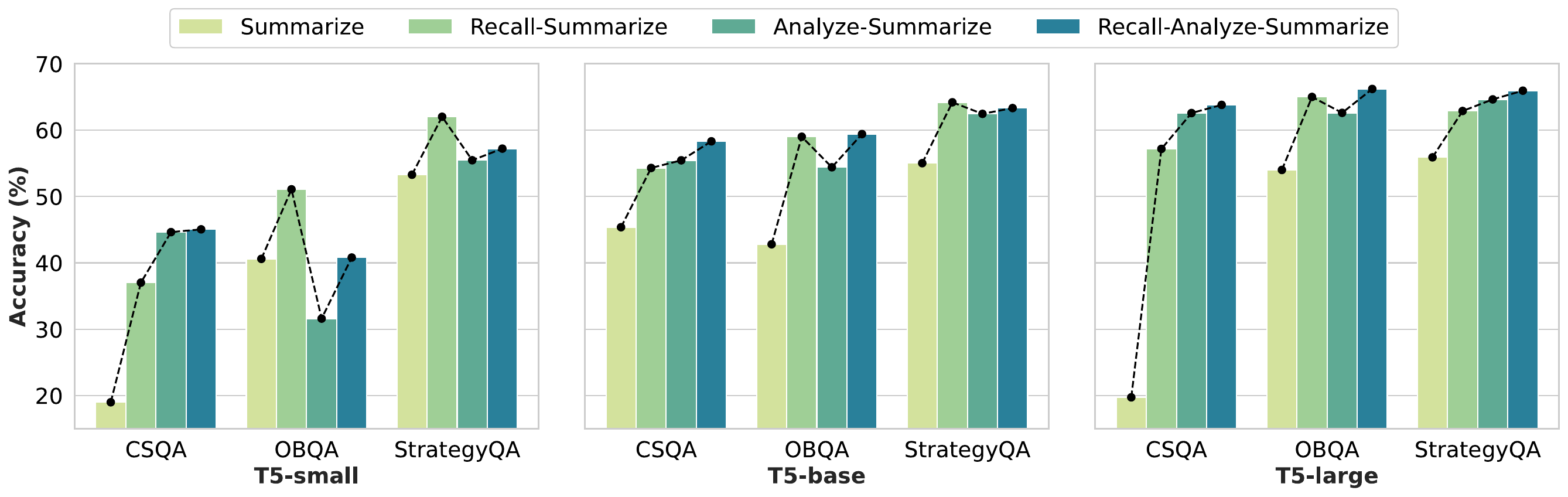}
  \caption {Ablation study on the effects of each stage in the recall-analyze-summarize reasoning process.}
  \label{fig:ablation-sft}
\end{figure*}

\begin{table*}[t]
\renewcommand{\arraystretch}{0.9}
    \centering
    \begin{tabular}{lccccc}
        \Xhline{1.0pt}
        \textbf{Model} & \textbf{Recall} & \textbf{Analyze} & \textbf{CSQA} & \textbf{OBQA} & \textbf{StrategyQA}\\ \Xhline{0.5pt}
        \multirow{4}{*}{\textit{T5-Small (60M)}}
        & \textcolor{red}{\ding{55}} & \textcolor{red}{\ding{55}} & 45.05 & 40.80 & 57.21\\ 
        & \textcolor{darkgreen}{\ding{51}} & \textcolor{red}{\ding{55}} & 45.54 \small{(+0.50)} & 45.20 \small{(+4.40)} & 58.08 \small{(+0.87)}\\ 
        & \textcolor{red}{\ding{55}} & \textcolor{darkgreen}{\ding{51}} & 46.02 \small{(+0.97)} & 50.00 \small{(+9.20)} & 59.39 \small{(+2.18)}\\ 
        & \textcolor{darkgreen}{\ding{51}} & \textcolor{darkgreen}{\ding{51}} & \textbf{46.36 \small{(+1.31)}} & \textbf{53.60 \small{(+12.80)}} & \textbf{60.26 \small{(+3.05)}}\\ 
        \Xhline{0.5pt}
        \multirow{4}{*}{\textit{T5-Base (220M)}}
        & \textcolor{red}{\ding{55}} & \textcolor{red}{\ding{55}} & 58.31 & 59.40 & 63.32\\ 
        & \textcolor{darkgreen}{\ding{51}} & \textcolor{red}{\ding{55}} & 58.72 \small{(+0.41)} & 60.19 \small{(+0.79)} & 64.63 \small{(+1.31)}\\ 
        & \textcolor{red}{\ding{55}} & \textcolor{darkgreen}{\ding{51}} & 59.54 \small{(+1.31)} & 61.80 \small{(+2.40)}& 65.07 \small{(+1.75)}\\ 
        & \textcolor{darkgreen}{\ding{51}} & \textcolor{darkgreen}{\ding{51}} & \textbf{59.79 \small{(+1.48)}} & \textbf{62.40 \small{(+3.00)}} & \textbf{66.38 \small{(+3.06)}}\\ 
        \Xhline{0.5pt}
        \multirow{4}{*}{\textit{T5-Large (770M)}}
        & \textcolor{red}{\ding{55}} & \textcolor{red}{\ding{55}} & 63.80 & 66.20 & 65.94\\ 
        & \textcolor{darkgreen}{\ding{51}} & \textcolor{red}{\ding{55}} & 64.13 \small{(+0.33)} & 66.80 \small{(+0.60)} & 65.94 \small{(+0.00)}\\ 
        & \textcolor{red}{\ding{55}} & \textcolor{darkgreen}{\ding{51}} & 65.19 \small{(+1.39)} & 67.40 \small{(+1.20)} & 68.12 \small{(+2.18)}\\ 
        & \textcolor{darkgreen}{\ding{51}} & \textcolor{darkgreen}{\ding{51}} & \textbf{65.44 \small{(+1.64)}} & \textbf{68.80 \small{(+2.60)}} & \textbf{69.00 \small{(+3.06)}}\\
        \Xhline{1.0pt}
    \end{tabular}
    \caption{Ablation study on the effect of applying DPO at recall and analyze stages. \textcolor{darkgreen}{\ding{51}} and \textcolor{red}{\ding{55}} indicate whether DPO is applied. All \textcolor{red}{\ding{55}} symbols for both stages denote performance after the reasoning acquisition phase, whereas individual \textcolor{darkgreen}{\ding{51}} symbols correspond to the recall-DPO and analyze-DPO setting, respectively. \textbf{Bold} values indicate the best performance, and values in parentheses represent the performance change after applying DPO in each setting.}
    \label{tab:ablation-dpo}
\end{table*}

\subsection{Ablation Study}
\paragraph{The effect of three-stage process} TinyThinker employs a structured reasoning process to progressively refine the student's knowledge. To evaluate the necessity of this stage-wise refinement, we conducted the following experiments: In \textbf{Summarize}, the student directly derives the summary from the question and options; in \textbf{Recall-Summarize}, the model first generates general knowledge during the recall stage, which is then used alongside the question and options to infer summary; in \textbf{Analyze-Summarize}, the model first generates specific knowledge during the analyze stage and combines it with the question and options to reach a summary.

As indicated in Figure~\ref{fig:ablation-sft}, performance generally improves with an increasing amount of available knowledge, particularly on the CSQA dataset, which demonstrates the efficacy of the reasoning process in progressively enhancing knowledge from coarse to fine granularity. Conversely, for the OBQA and StrategyQA datasets, the optimal performance for the T5-small and T5-base models was achieved using only general knowledge. This can be attributed to the model's parameter size, which limits its capacity to manage the increasing complexity of later stages, particularly during the analyze and summarize stages. While generating general knowledge requires a broad understanding of the question and options, generating specific knowledge requires a more detailed assessment of each option,significantly increasing the computational complexity of the later stages.

Consequently, as model size increases, the performance gap between Recall-Summarize and Analyze-Summarize gradually narrows, leading to improved performance on Recall-Analyze-Summarize. This trend is observable across both the OBQA and StrategyQA datasets.

\paragraph{The effect of self-reflection} After training the student model to reason through the three-stage process, we applied the DPO algorithm to both the recall and analyze stages to further consolidate its reasoning capabilities. This approach has two key objectives: reinforcing correct knowledge and refining incorrect knowledge. In this experiment, we investigate the performance gains of applying DPO individually to the recall and analyze stages, denoted as \textbf{recall-DPO} and \textbf{analyze-DPO}, respectively.

As shown in Table~\ref{tab:ablation-dpo}, applying DPO yielded performance improvements at both stages, with the most substantial gains observed when DPO was applied concurrently to both the recall and analyze stages. Notably, the improvement in the analyze stage was more pronounced than in the recall stage, supporting our earlier observation that the analyze stage is inherently more challenging to learn. This suggests that further performance gains could be achieved by refining learning strategies or increasing model size to better handle the complexities of the analyze stage.

\section{Conclusion}
In this study, we introduced TinyThinker to enhance reasoning capabilities through effective knowledge internalization. In contrast to prior methods, we developed a structured three-stage reasoning process that progressively refines knowledge from coarse to fine granularity. This process is complemented by a two-phase training approach, consisting of reasoning acquisition and self-refinement phases. Experiments on commonsense reasoning benchmarks demonstrate that TinyThinker achieves superior performance compared to existing baselines. Additionally, the ablation study further confirms the contributions of each component, highlighting the overall effectiveness of our approach. We expect that TinyThinker provides a flexible framework extendable to other knowledge-intensive reasoning tasks, offering a promising strategy for developing effective reasoning capabilities in smaller LMs.

\section{Limitation}
\paragraph{Quality of curated data} Although LLMs are capable of generating semantically coherent sentences, their inherent issue of hallucination sometimes leads to content that lacks factual accuracy and safety \citep{ji2023survey}. This lack of factual accuracy also affects reasoning-related content generation, particularly through the "error cascade" problem, where a factual error in an intermediate reasoning step propagates inaccuracies through subsequent steps \citep{chu-etal-2024-navigate}. Therefore, ensuring the factual quality of the data generated by the teacher model remains a challenges in this study.

\paragraph{Efficient generation strategy} While the proposed three-stage reasoning process is effective for multiple-choice tasks, there is room for improving efficiency, especially in the analyze stage. Currently, the model generates specific knowledge independently for each option, leading to a time-intensive process. A more efficient approach would generate specific knowledge for all options in parallel, thereby accelerating the analyze stage.

\section*{Acknowledgments}
This research was supported by the National Research Foundation (NRF) funded by the Korean government (MSIT) (No. RS-2023-00229822). We sincerely thank Shengen Piao, Jieun Lee, and Huijun Jin for their constructive suggestions and discussions, which significantly contributed to improving this work.

\bibliography{custom}

\appendix

\section{Hyperparameters}
\label{sec:hyperparameter}
The hyperparameters utilized in this study are summarized in Table~\ref{tab:hyperparameter}.

\section{Datasets}
\label{sec:datasets}
\paragraph{CSQA} CSQA is a multiple-choice question-answering dataset with five answer options per question, which requires diverse commonsense knowledge. Since the test set is not publicly available, we report performance on the validation set, following \citet{li2024explanations, wang-etal-2023-scott, hsieh-etal-2023-distilling}. Additionally, we randomly sampled 1,221 questions from the training set to form a development set, resulting in a final split of 8,520/1,221/1,221 questions for the training/validation/test sets.

\paragraph{OBQA} OBQA is a four-choice question-answering dataset designed to evaluate the ability to apply broad common knowledge, particularly for elementary-level science questions. The dataset contains 4,957/500/500 questions for the training/validation/test sets.

\paragraph{StrategyQA} StrategyQA is a binary (yes/no) question-answering dataset that requires implicit reasoning across diverse topics. The training set contains 2,290 questions, while the test set includes 490 questions. As the test set is not publicly available, we split the training set into 80\% for training, 10\% for validation, and 10\% for testing, following the procedure outlined in \citet{magister-etal-2023-teaching} to ensure reproducibility.

To enhance the diversity of teacher-generated data during data curation, we utilize temperature sampling to generate 4-8 distinct reasoning processes per question to ensure diversity in the generated data. As demonstrated by SCoTD \citep{li-etal-2023-symbolic}, diverse reasoning data is crucial for effective reasoning distillation. After generation, we apply filtering procedures, such as de-duplication, to ensure data quality. The statistics of the curated datasets are summarized in Table~\ref{tab:curated_dataset}.

\section{Full Prompts}
\label{sec:full prompts}
The prompt design follows a consistent instruction template across all datasets, as illustrated in Table~\ref{tab:instruction}. This instruction provides foundational guidance for the teacher model, facilitating effective data generation. Subsequently, few-shot examples for each dataset are listed in Table~\ref{tab:prompt_csqa}, Table~\ref{tab:prompt_obqa}, and Table~\ref{tab:prompt_strategyqa}.

\onecolumn
\begin{table}
    \centering
    \begin{tabular}{ccl}
        \Xhline{1.0pt}
        \textbf{Hyperparameter} & \textbf{Value} & \textbf{Note}\\ \Xhline{0.5pt}
        \multicolumn{3}{c}{\textbf{\textit{Data Curation}}} \\
        max\_tokens & 256 & Maximum tokens generated per input. \\
        n & [4, 8] & Number of generations for each input message. \\
        temperature & 0.8 & Temperature value for sampling. \\ \Xhline{0.5pt}
        \multicolumn{3}{c}{\textbf{\textit{Reasoning Acquisition}}} \\
        epochs & 10 & Total number of training epochs.\\
        batch\_size & 64 & Batch size for training.\\
        interval & 100 & Default number of steps for recall, analyze, summarize stage.\\
        lr & $5\times10^{-4}$ & Learning rate of AdamW optimizer.\\ \Xhline{0.5pt}
        \multicolumn{3}{c}{\textbf{\textit{Self-Reflection}}} \\
        iterations & 5 & Number of iterations for DPO training.\\
        n & 10 & Generation number for pairwise data collection. \\
        temperature & 0.7 & Temperature value for sampling. \\
        epochs & 10 & Epochs for each DPO iteration.\\
        batch\_size & 64 & Batch size for DPO training.\\
        lr & $5\times10^{-6}$ & Learning rate of AdamW optimizer.\\
        $\beta$ & 0.5 & Weight of beta value. \\
        $\alpha$ & 0.5 & Weight of NLL lose. \\
        \Xhline{1.0pt}
    \end{tabular}
    \caption{Hyperparameter settings used in the experiments.}
    \label{tab:hyperparameter}
\end{table}

\begin{table}
    \centering
    \begin{tabular}{lccccc}
        \Xhline{1.0pt}
        \textbf{Dataset} & \textbf{\# Recall} & \textbf{\# Analyze} & \textbf{\# Summarize}& \textbf{\# Validation} & \textbf{\# Test} \\ \Xhline{0.5pt}
        CSQA & 66,739 & 334,331 & 66,869 & 1,221 & 1,221 \\
        OBQA & 19,444 & 77,985 & 19,497 & 500 & 500 \\
        StrategyQA &7,089 & 14,290 & 7,145 & 229 & 229 \\
        \Xhline{1.0pt}
    \end{tabular}
    \caption{Dataset statistics following data curation.}
    \label{tab:curated_dataset}
\end{table}

\begin{tcolorbox}[colback=gray!10!white, colframe=black, title=\textbf{\textsc{Instruction}}, width=\linewidth, boxrule=1pt]
Reference examples delimited with \#\#\# as a guide to outline the reasoning steps towards the correct answer in a step-by-step manner.
First, concisely summarize the key information or core issue from the question statement. Then, systematically evaluate each option according to the following format:\newline\newline
Key Information: [Summarize the key information from the question.] \newline
Explanations: [Option] is [correct/incorrect]. Because [Explain the option's features and relevance to the key information.]\newline\newline
Ensure explanations are clear, concise, and diverse by using varied logic, examples, and perspectives. Be creative to prevent repetition and make sure each step leads to the summary. Highlighting the uniqueness of each example to support diverse reasoning.
\end{tcolorbox}
\begin{minipage}{\textwidth}
\captionof{table}{The instruction of data curation.}
\label{tab:instruction}
\end{minipage}

\begin{tcolorbox}[colback=gray!10!white, colframe=black, title=\textbf{\textsc{Prompt for CSQA}}, width=\textwidth, boxrule=1pt, breakable]
\textbf{1.} The fox walked from the city into the forest, what was it looking for?\newline
Options: (A) pretty flowers (B) hen house (C) natural habitat (D) storybook\newline
Key Information: The motivations of a fox moving from an urban to a forest environment, focusing on natural instincts or necessities.\newline
Explanations: A is incorrect. Because Foxes seeking pretty flowers does not align with their instinctual needs for survival, such as food or shelter.\newline
B is incorrect. Because a hen house, usually near human habitats, does not match the forest setting the fox is moving towards.\newline
C is correct. Because the fox's migration to the forest implies a search for its natural habitat, indicating a pursuit of basic needs and instincts.\newline
D is incorrect. Because a storybook, as a human creation, doesn't meet any of a fox's natural survival instincts in a forest setting.
\newline\newline
\textbf{2.} Sammy wanted to go to where the people were. Where might he go?\newline
Options: (A) populated areas (B) race track (C) desert (D) apartment (E) roadblock\newline
Key Information: Sammy seeks location that is likely to be frequented by or filled with individuals.\newline
Explanations: A is correct. Because populated areas naturally have many people, aligning with Sammy's goal.\newline
B is incorrect. Because a race track may be crowded during events but doesn't consistently draw people, not fully aligning with Sammy's aim.\newline
C is incorrect. Because deserts are sparsely populated and do not meet Sammy's desire to be around people.\newline
D is incorrect. Because while an apartment indicates residence, it doesn't broadly represent a gathering place for many people.\newline
E is incorrect. Because roadblocks might temporarily gather people but aren't places people intentionally seek for socializing.
\newline\newline
\textbf{3.} What do people use to absorb extra ink from a fountain pen?\newline
Options: (A) shirt pocket (B) calligrapher's hand (C) inkwell (D) desk drawer (E) blotter\newline
Key Information: People use a specific tool or material to absorb extra ink from a fountain pen.\newline
Explanations: A is incorrect. Because a shirt pocket may catch ink accidentally but is not used intentionally for absorbing extra ink from a fountain pen.\newline
B is incorrect. Because a calligrapher's hand might contact with ink but is not used purposefully to absorb extra ink.\newline
C is incorrect. Because an inkwell is for storing ink, not absorbing excess ink from a pen.\newline
D is incorrect. Because a desk drawer is for storage, not specifically for absorbing excess ink.\newline
E is correct. Because a blotter is designed to absorb excess ink, preventing smudges and keeping the writing area clean.
\newline\newline
\textbf{4.} Before getting a divorce, what did the wife feel who was doing all the work?\newline
Options: (A) harder (B) anguish (C) bitterness (D) tears (E) sadness\newline
Key Information: The emotional state of a wife who perceived imbalance of responsibilities before deciding on divorce, suggesting feelings stemming from stress, imbalance, or discontent.\newline
Explanations: A is incorrect. Because 'harder' refers to the level of effort, not an emotional state, and doesn't match the emotional context of imbalance leading to divorce.\newline
B is incorrect. Because 'anguish' indicates severe distress, it may not fully capture the specific feelings of imbalance and discontent in this scenario.\newline
C is correct. Because 'bitterness' accurately reflects feelings of anger and disappointment from unmet expectations and perceived imbalance, aligning with the scenario.\newline
D is incorrect. Because 'tears' indicate a physical manifestation of emotions but do not specify the emotional state related to the feeling of being overburdened.\newline
E is incorrect. Because 'sadness' is a general emotion that doesn't precisely convey the sense of being undervalued or the specific discontent that led to considering divorce.
\newline\newline
\textbf{5.} Where do you put your grapes just before checking out?\newline
Options: (A) mouth (B) grocery cart (C) super market, (D) fruit basket (E) fruit market\newline
Key Information: One would place grapes in the immediate moments before proceeding to the checkout in a shopping context, as part of the  grocery purchasing process.\newline
Explanations: A is incorrect. Because putting grapes in one's mouth before checkout implies eating them before purchase, which is inappropriate.\newline
B is correct. Because the grocery cart is used for holding items to be purchased, right before checking out.\newline
C is incorrect. Because the supermarket is the overall shopping location, not where one places items immediately before checkout.\newline
D is incorrect. Because a fruit basket is  typically for storing grapes at home, not for holding them just before checkout.\newline
E is incorrect. Because the fruit market refers to the broader shopping venue, not the specific spot for grapes just before checkout.
\newline\newline
\textbf{6.} Google Maps and other highway and street GPS services have replaced what?\newline
Options: (A) United States (B) Mexico (C) countryside (D) atlas\newline
Key Information: There is a transition from traditional navigation tools to modern navigation tools, focusing on what has been predominantly replaced by digital mapping and GPS services in terms of functionality.\newline
Explanations: A is incorrect. Because the United States, being a country, cannot be replaced by digital mapping technologies.\newline
B is incorrect. Because Mexico, as a country, cannot be replaced by digital mapping technologies, highlighting an incorrect match with navigation tools.\newline
C is incorrect. Because the countryside, a type of geographic area, cannot be replaced by GPS services in the context of navigation tool replacement.\newline
D is correct. Because an atlas, a book of maps, has been functionally replaced by digital mapping and GPS services, marking a shift in how people navigate.
\newline\newline
\textbf{7.} What home entertainment equipment requires cable?\newline
Options: (A) radio shack (B) substation (C) television (D) cabinet\newline
Key Information: Devices used for entertainment purposes within a home setting that necessitate a connection through a physical cable for operation or functionality\newline
Explanations: A is incorrect. Because radio shack refers to a place rather than a piece of home entertainment equipment requiring a cable.\newline
B is incorrect. Because a substation pertains to electrical power distribution, not directly related to home entertainment or devices needing a cable in a home.\newline
C is correct. Because televisions typically require a cable connection for signal reception, aligning with the key concept of home entertainment equipment needing a cable.\newline
D is incorrect. Because a cabinet is used for storage and does not require a cable for home entertainment purposes, unrelated to the key concept.
\newline\newline
\textbf{8.} The man laid on the soft moss and looked up at the trees, where was the man?\newline
Options: (A) Niagra Falls (B) forest (C) waterfall (D) ground (E) tree\newline
Key Information: The scenario describes a natural setting characterized by soft moss and an upward view of trees, suggesting a location where these features are prominent.\newline
Explanations: A is incorrect. Because Niagara Falls is primarily associated with its waterfalls, not a setting of soft moss and tree views.\newline
B is correct. Because a forest offers both soft moss to lie on and numerous trees to look up at, matching the scenario's description.\newline
C is incorrect. Because while waterfalls can be found in forests, the specific mention of laying on soft moss and looking up at trees directly suggests a broader environment than just the area near a waterfall.\newline
D is incorrect. Because while the man is technically on the ground, it is too general and misses the specific natural characteristics implied by the details of soft moss and trees.\newline
E is incorrect. Because the scenario describes  the man is looking up at trees, not situated in one, conflicting with the scenario given.\\
\end{tcolorbox}
\begin{minipage}{\textwidth}
\captionof{table}{The prompt for CSQA dataset.}
\label{tab:prompt_csqa}
\end{minipage}

\begin{tcolorbox}[colback=gray!10!white, colframe=black, title=\textbf{\textsc{Prompt for OBQA}}, width=\textwidth, boxrule=1pt, breakable]
\textbf{1.} As a car approaches you in the night,\newline
Options: (A) the headlights become more intense (B) the headlights recede into the dark (C) the headlights remain at a constant (D) the headlights turn off\newline
Key Information: The intensity of headlights appears to increase as a car approaches due to the reduction in distance, allowing more light to reach the observer.\newline
Explanations: A is correct. Because the intensity of headlights increases as the distance decreases, consistent with the concept of light appearing more intense as a car approaches.\newline
B is incorrect. Because headlights become more visible, not less, as a car approaches, contradicting the concept of increasing intensity with proximity.\newline
C is incorrect. Because the apparent brightness of headlights increases as the car comes closer, not remaining constant, diverging from the notion of constant intensity.\newline
D is incorrect. Because headlights typically do not turn off as a car approaches, unrelated to the concept of light behavior with respect to distance.
\newline\newline
\textbf{2.} What is the most likely to be an effect of acid rain on an aquatic environment?\newline
Options: (A) decrease in plant life (B) increase in fish population (C) increase in plant growth (D) cleaner and clearer water\newline
Key Information: Acid rain, containing harmful acidic compounds, detrimentally affects aquatic environments by altering water chemistry, harmful to aquatic life.\newline
Explanations: A is correct. Because acid rain reduces water pH, harming aquatic plants, consistent with the concept that acid rain negatively affects aquatic life.\newline
B is incorrect. Because acid rain decreases fish populations by introducing harmful chemicals, contrary to the idea of an increase.\newline
C is incorrect. Because acid rain impedes plant growth due to its acidity, opposing the concept of promoting a healthy aquatic environment.\newline
D is incorrect. Because acid rain pollutes water, reducing clarity, aligning with the concept of its harmful impacts on aquatic environments.
\newline\newline
\textbf{3.} The moon's surface\newline
Options: (A) is smooth on the entire surface (B) contains large cavities cause by explosions (C) contains an internal core of cheese (D) is filled with lakes\newline
Key Information: The moon's surface is characterized by a variety of geological features, including craters from asteroid impacts.\newline
Explanations: A is incorrect. Because the moon's surface is not smooth but covered in craters and rough terrain, contradicting the idea of uniform smoothness.\newline
B is correct. Because the moon's surface features large craters caused by asteroid impacts, aligning with the concept of its varied geological features.\newline
C is incorrect. Because the moon is made of rock, not cheese, contrasting with scientific evidence of its geological composition.\newline
D is incorrect. Because the moon lacks lakes or liquid water, diverging from the key concept by misrepresenting its surface conditions.
\newline\newline
\textbf{4.} When the weather changes as it does from Christmas to Easter,\newline
Options: (A) the air may chill (B) the ground may freeze (C) the plants may die (D) the ground may warm\newline
Key Information: The period from Christmas to Easter typically involves a transition from winter to spring in many parts of the world, marked by gradually increasing temperatures and thawing conditions.\newline
Explanations: A is incorrect. Because the trend from Christmas to Easter is toward warmer weather, not colder, contrasting with the concept of seasonal warming.\newline
B is incorrect. Because ground freezing aligns more with winter's onset; the period to Easter typically sees thawing, opposing the warming trend concept.\newline
C is incorrect. Because the approach to Easter, signaling spring, is a time for plant life to begin thriving again, not dying, contradicting the concept of seasonal renewal.\newline
D is correct. Because the ground warming from Christmas to Easter reflects the transition from winter to spring, consistent with the concept of increasing temperatures.
\newline\newline
\textbf{5.} Heat and moisture in the ocean is a good recipe for\newline
Options: (A) a violent storm (B) violent sea animals (C) condensation (D) inland storms\newline
Key Information: Heat and moisture in the ocean contribute to the formation of storms by providing the energy and water vapor necessary for their development.\newline
Explanations: A is correct. Because heat and moisture are key ingredients for storm development over the ocean, directly aligning with the concept of storm formation.\newline
B is incorrect. Because the link between heat and moisture and 'violent sea animals' is not relevant to the meteorological focus of the key concept.\newline
C is incorrect. Because while condensation is a component of storm formation, it does not fully encompass the broader process of storm development indicated by the key concept.\newline
D is incorrect. Because heat and moisture contribute to storms that can affect inland areas, demonstrating the widespread impact of these conditions beyond the immediate ocean environment.
\newline\newline
\textbf{6.} Hummingbirds take what with them\newline
Options: (A) Bees (B) energy (C) Pollen (D) Honey\newline
Key Information: As pollinators, hummingbirds interact with flowers for nectar, during which they inadvertently collect and transfer pollen.\newline
Explanations: A is incorrect. Because hummingbirds do not transport bees, unrelated to their role as pollinators.\newline
B is incorrect. Because while hummingbirds expend energy, saying they 'take energy' is too abstract and unrelated to their pollination activities.\newline
C is correct. Because pollen is transferred by hummingbirds as they feed on nectar, aligning with their role as pollinators.\newline
D is incorrect. Because hummingbirds consume nectar, not honey, which is unrelated to their interaction with flowers
\newline\newline
\textbf{7.} What covers over 90\% of the Earth's surface and 0\% of the moon's surface\newline
Options: (A) a magnesium iron silicate mineral (B) chemical element with the symbol S (C) the element with the symbol Fe (D) that which contains 2 hydrogen and 1 oxygen molecules\newline
Key Information: A substance that is abundant on Earth's surface but nonexistent on the moon's surface, pointing towards Earth's unique characteristic related to its surface coverage.\newline
Explanations: A is incorrect. Because magnesium iron silicate minerals are primarily found in Earth's mantle, not its surface, and have no relevance to the moon's surface.\newline
B is incorrect. Because Sulfur does not cover Earth's surface, making it irrelevant to the comparison with the moon.\newline
C is incorrect. Because iron (Fe) is common in both Earth's and the moon's compositions but does not cover Earth's surface in the context required by the question.\newline
D is correct. Because water, composed of 2 hydrogen and 1 oxygen molecules (H2O), covers over 70\% of Earth's surface but 0\% of the moon's surface, aligning with Earth's unique surface characteristics.\\
\end{tcolorbox}
\begin{minipage}{\textwidth}
\captionof{table}{The prompt for OBQA dataset.}
\label{tab:prompt_obqa}
\end{minipage}

\begin{tcolorbox}[colback=gray!10!white, colframe=black, title=\textbf{\textsc{Prompt for StrategyQA}}, width=\textwidth, boxrule=1pt, breakable]
\textbf{1.} Do hamsters provide food for any animals?\newline
(A) yes (B) no\newline
Key Information: Hamsters serve as prey for other animals in the food chain.\newline
Explanations: A is correct. Because hamsters are prey for various predators, reflecting their role in the food chain and ecosystem interconnectedness.\newline
B is incorrect. Because denying this ignores the reality of food chain dynamics and hamsters' role in supporting habitat biodiversity.
\newline\newline
\textbf{2.} Could Brooke Shields succeed at University of Pennsylvania?\newline
(A) yes (B) no\newline
Key Information: The potential for success of an individual, Brooke Shields, at a specific academic institution, University of Pennsylvania, implying considerations of her capability, ambition, and the university's environment.\newline
Explanations: A is correct. Because Brooke Shields' academic and career accomplishments suggest she would likely succeed at the University of Pennsylvania, reflecting her capability to thrive in demanding environments.\newline
B is incorrect. Because there's ample evidence of Brooke Shields' ability and ambition, making her success at the University of Pennsylvania plausible, thus dismissing this ignores her proven capability.
\newline\newline
\textbf{3.} Hydrogen's atomic number squared exceeds number of Spice Girls?\newline
(A) yes (B) no\newline
Key Information: Hydrogen's atomic number is 1, and the Spice Girls group consists of 5 members.\newline
Explanations: A is incorrect. Because hydrogen's atomic number squared (1) does not exceed the Spice Girls' member count (5).\newline
B is correct. Because since hydrogen's atomic number squared equals 1, it is less than the Spice Girls' 5 members, aligning with the key concept.
\newline\newline
\textbf{4.} Is it common to see frost during some college commencements?\newline
(A) yes (B) no\newline
Key Information: College commencements in certain regions or during certain times of the year might coincide with colder weather conditions, which can lead to the formation of frost.\newline
Explanations: A is correct. Because frost can occur during college commencements in colder climates or seasonal transitions, reflecting geographic and seasonal weather variations.\newline
B is incorrect. Because disregarding frost ignores the reality of seasonal cold weather affecting commencements in many regions.
\newline\newline
\textbf{5.} Could a llama birth twice during War in Vietnam (1945-46)?\newline
(A) yes (B) no\newline
Key Information: The natural gestation period of a llama is approximately 11 months, and the specific duration of the War in Vietnam stated to be 6 months.\newline
Explanations: A is incorrect. Because the llama's 11-month gestation period surpasses the 6-month Vietnam War duration, making two births impossible within this time frame.\newline
B is correct. Because given the llama's gestation period and the war's duration, it's biologically impossible for a llama to give birth twice during the conflict.
\newline\newline
\textbf{6.} Would a pear sink in water?\newline
(A) yes (B) no\newline
Key Information: The density of pear would determine if it sinks or floats in water.\newline
Explanations: A is incorrect. Because pears float due to a density less than water, attributed to their fibrous and airy composition.\newline
B is correct. Because pears float in water because of their air-filled fibrous structure, making them less dense than water.
\newline\newline
\textbf{7.} Is Albany, Georgia the most populous US Albany?\newline
(A) yes (B) no\newline
Key Information: Comparing the population size between Albany, Georgia and Albany, New York.\newline
Explanations: A is incorrect. Because Albany, New York, has a larger population than Albany, Georgia, making the latter not the most populous.\newline
B is correct. Because Albany, New York, surpasses Albany, Georgia, in population, thereby making the Georgia city not the most populous Albany.
\newline\newline
\textbf{8.} Can the Great Depression be treated with Prozac?\newline
(A) yes (B) no\newline
Key Information: Great Depression is a historical event and Prozac is a medication used to treat clinical depression.\newline
Explanations: A is incorrect. Because Prozac treats clinical depression, not historical events like the Great Depression.\newline
B is correct. Because the Great Depression, an economic crisis, cannot be treated with Prozac, a medication for clinical depression. \\
\end{tcolorbox}
\begin{minipage}{\textwidth}
\captionof{table}{The prompt for StrategyQA dataset.}
\label{tab:prompt_strategyqa}
\end{minipage}

\end{document}